\documentclass[10pt,twocolumn,letterpaper]{article}

\usepackage{iccv}
\usepackage{times}
\usepackage{epsfig}
\usepackage{graphicx}
\usepackage{amsmath}
\usepackage{amssymb}

\usepackage{soul}
\usepackage{bbding}
\usepackage{pifont}
\usepackage{wasysym}

\newcommand{\cmark}{\ding{52}}%
\newcommand{\xmark}{\ding{56}}%

\usepackage[pagebackref=true,breaklinks=true,letterpaper=true,colorlinks,bookmarks=false]{hyperref}

\iccvfinalcopy % *** Uncomment this line for the final submission

\ificcvfinal\fi
\begin{document}

%%%%%%%%% TITLE
\title{PathGAN: Local Path Planning with Attentive Generative Adversarial Networks}

\author{Dooseop Choi, Seung-Jun Han, Kyoungwook Min, Jeongdan Choi\\
ETRI\\
Republic of Korea\\
{\tt d1024.choi@etri.re.kr}
}

\maketitle
%\thispagestyle{empty}

%%%%%%%%% ABSTRACT
\begin{abstract}
    To achieve autonomous driving without high-definition maps, we present a model capable of generating multiple plausible paths from egocentric images for autonomous vehicles. Our generative model comprises two neural networks: the feature extraction network (FEN) and path generation network (PGN). The FEN extracts meaningful features from an egocentric image, whereas the PGN generates multiple paths from the features, given a driving intention and speed. To ensure that the paths generated are plausible and consistent with the intention, we introduce an attentive discriminator and train it with the PGN under generative adversarial networks framework. We also devise an interaction model between the positions in the paths and the intentions hidden in the positions and design a novel PGN architecture that reflects the interaction model, resulting in the improvement of the accuracy and diversity of the generated paths. Finally, we introduce ETRIDriving, a dataset for autonomous driving in which the recorded sensor data are labeled with discrete high-level driving actions, and demonstrate the state-of-the-art performance of the proposed model on ETRIDriving in terms of accuracy and diversity.
\end{abstract}

%%%%%%%%% BODY TEXT
\section{Introduction}
In recent years, autonomous driving has witnessed significant advances owing to breakthroughs in sensor technology and artificial intelligence. A few companies are providing commercial autonomous vehicle (AV) services in a limited area, and others are testing their AVs with the goal of commercialization. Most automated driving systems (ADSs) for AVs are known to rely on an algorithmic framework, which comprises three stages: $\textit{perception}$, $\textit{planning}$, and $\textit{control}$, based on a high-definition (HD) map. In the perception stage, the system recognizes the static and dynamic objects surrounding an AV. Based on the recognition, it decides the next movements of the AV and controls it in the planning and control stages, respectively. 

HD maps enable AVs to see beyond the coverage of the mounted sensors by providing an accurate representation of the road ahead and information on the surrounding environment~\cite{Katra}. Furthermore, the HD map data ensure the precise localization of AVs, with the surrounding lane lines or landmarks as reference positions. However, the high dependence of AVs on HD maps sometimes limits their ability to drive in diverse driving environments. Assuming a vehicle is on an unpaved road or a road with recently repainted lane lines, human drivers can determine plausible paths for the vehicle based on the contextual cues of the surroundings (e.g., the lane lines, shape of the road, and locations of the static and dynamic objects). The conventional ADSs, however, may have difficulty in determining the paths, if the HD map does not provide the cues.

One possible solution is to build a system that can directly extract the contextual cues from sensor data (images and point clouds) and use them to control the vehicles. Many approaches have been proposed in the literature to generate control signals (such as steering angle and speed) from the sensor data~\cite{Pomerleau, Bojarski, Xu, Codevilla, Hecker, Yang, Codevilla2, Buhet, Malla, kim3}. However, the existing methods have at least one of the following limitations: 1) they only generate control signals for one or a few maneuvers ($\textit{Go}$, $\textit{Turn left/right}$); 2) only one control signal is generated at a time; 3) the generated control signals are not interpretable.

\begin{figure}
\centerline {\includegraphics[width=9.0cm]{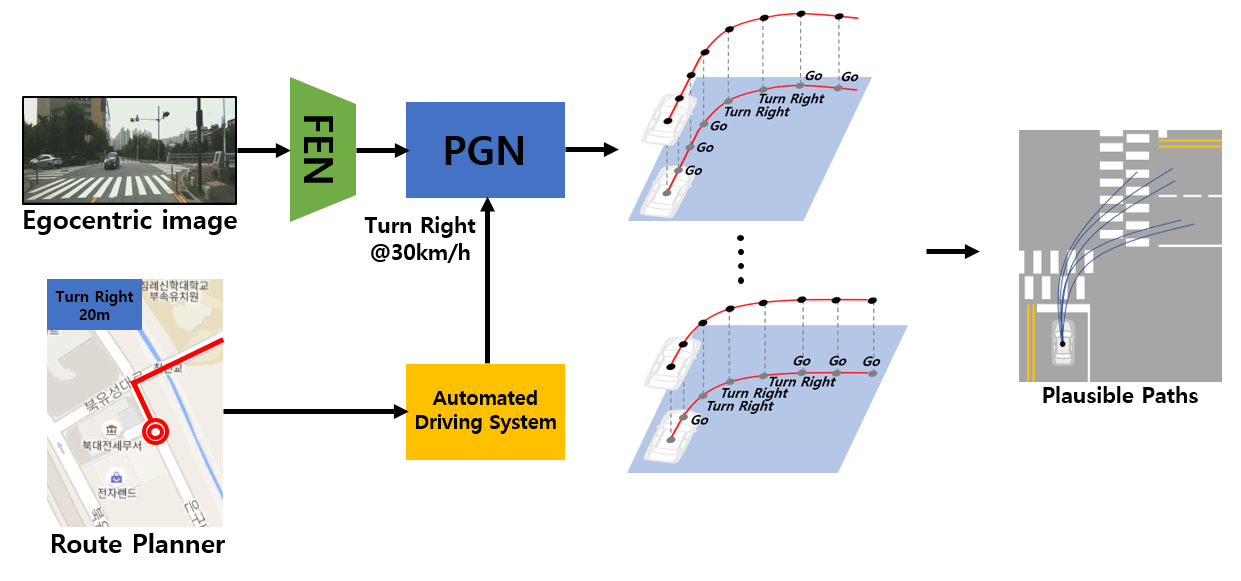}}
\caption{Overview of our path generation model. Our model generates plausible multiple paths consistent with input driving intentions and speed from egocentric images via a novel interaction model between the positions in the paths and the intentions hidden in the positions. }
\label{fig:overall}
\end{figure}

To address aforementioned limitations, we propose a model capable of generating plausible multiple paths consistent with the input driving intentions. The paths are generated based on the egocentric coordinate system. Therefore, the ADS can assess the risk of driving along a path at a target speed (interpretable) and use the path to calculate the control signals. Figure \ref{fig:overall} illustrates the overview of the proposed model. To the best of our knowledge, our model is the first that meets all of the following properties: 1) the model generates multiple paths corresponding to nine different maneuvers, 2) to generate paths consistent with the nine maneuvers, only one NN model is trained, 3) geometrical information (e.g., HD map, goal position) is not required for the path generation. As verified in sec. 5, simply modifying existing models \cite{Lee, Sadeghian2} to take a target maneuver as an input does not work well. 

The currently available autonomous driving datasets~\cite{Geiger, Xu, Hecker, Argoverse, Nuscene} are not suitable for training path generation models with the aforementioned characteristics because the driving data provided are not labeled with high-level driving actions of the ego-vehicle. For the task at hand, we classified the possible actions of a vehicle into nine types ($\textit{Go}$, $\textit{Turn Left/Right}$, $\textit{U-Turn}$, $\textit{Left/Right Lane Change}$, $\textit{Avoidance}$, $\textit{Left/Right Way}$) and manually assigned one of the nine actions to each frame of the recorded data that we collected while driving in diverse driving conditions. We refer to our dataset, which we introduce in sec. 3, as ETRIDriving. 
% figure 1 --------------------------------------------------------------------
% figure 1 --------------------------------------------------------------------
\begin{figure}
\centerline {\includegraphics[width=9.0cm]{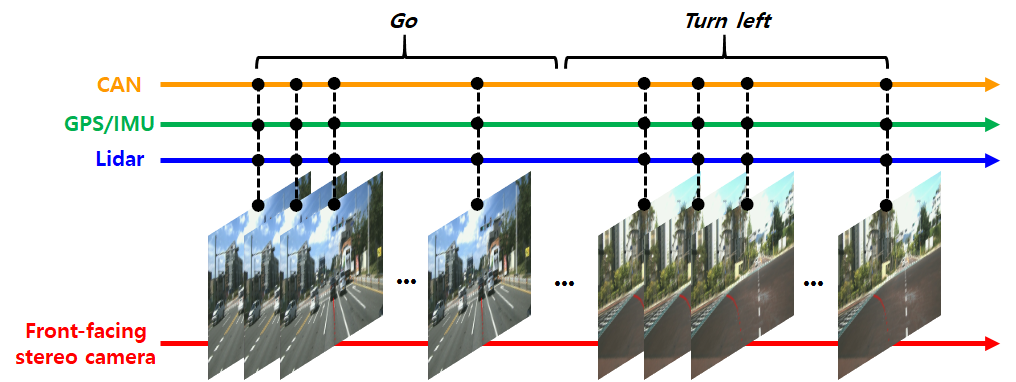}}
\caption{Brief description of our driving-action labeling process.}
\label{fig:one}
\end{figure}

\section{Related Works}
\textbf{End-to-End Models} Approaches in this category train NNs to map sensor data to control signals (steering angle and speed). The first attempt to exploit deep convolutional neural networks (CNNs) for the mapping was made by \cite{Bojarski}, who trained a CNN to map a front-facing camera image to a steering angle. Motivated by \cite{Bojarski}, many end-to-end models have been proposed in the literature. As mentioned in sec. 1, however, the models ignore two requirements that are important to real-world driving: the multimodal nature of a human driver's actions and the interpretability of the generated control signals.

\textbf{Future Trajectory Forecasting} Trajectory forecasting has received considerable attention in autonomous driving because predicting the movements of surrounding objects is essential for safe driving. In general, approaches in this category train recurrent neural networks (RNNs) to predict the trajectories of agents in a scene based on the scene context information available from the sensor data~\cite{Alahi, Lee, Gupta, Sadeghian2, Sadeghian, Rhinehart, Rhinehart2, Zeng, Choi, Fang}. Our model is closely related to the forecasting models in terms of 1) predicting multiple trajectories and 2) utilizing sensor data for understanding scenes. However, our model has, at least, one of the following advantages over these models: 1) it can generate multiple paths that match a driving intention from sensor data; 2) it does not require HD maps for path generation. Recently, \cite{Cai} proposed a model that estimates the future trajectory of an ego-vehicle from egocentric images, given a driving intention. However, multiple NN models of the same structure, each corresponding to a different driving intention, must be trained separately.

\textbf{Image Captioning} Image captioning is the process of generating textual descriptions of an image. Many approaches have been proposed in the literature~\cite{Xu2, Wu, Lu, Yao, Guo, Yang2}. These approaches are generally based on the encoder-decoder architecture in which a CNN extracts features from an image and RNNs generate the textual descriptions from the features. Motivated by the captioning models, we designed our path generation model. Specifically, our model is closely related to \cite{Aneja}, in which the intentions hidden in the generated words are captured in a sequential latent space through a variational autoencoder (VAE). Our approach is different from that of \cite{Aneja} in the following aspects: 1) we train our model to estimate the intentions hidden in the positions in the paths directly and learn the distribution of the sequential intentions at the same time through generative adversarial networks (GAN) framework \cite{Goodfellow}, 2) we design our model to utilize the visual contexts effectively for both the path generation and intention estimation through a novel NN architecture.

% figure 2 --------------------------------------------------------------------
\begin{figure*}
\centerline {\includegraphics[width=18.0cm]{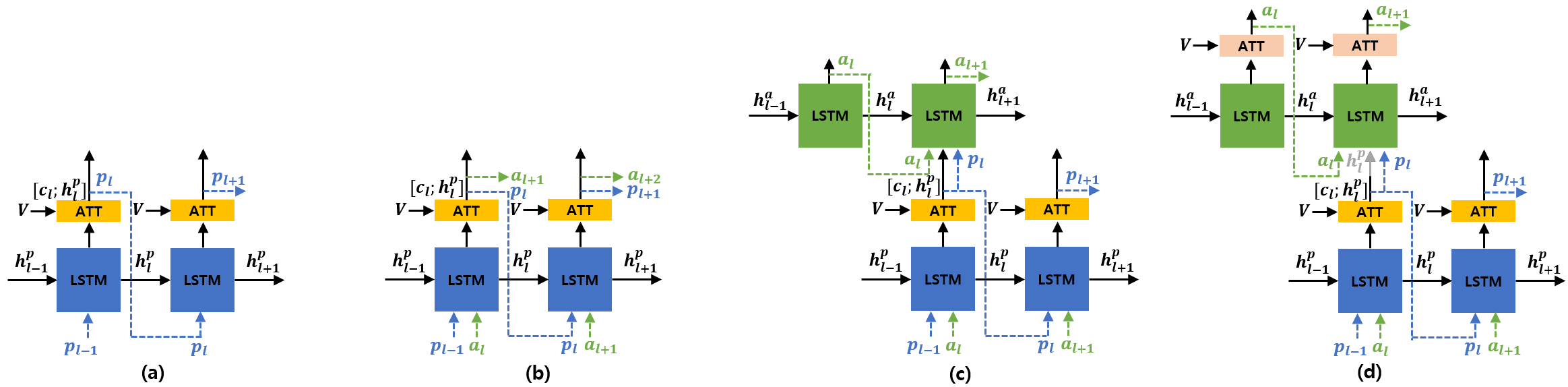}}
\caption{Brief descriptions of path generation networks.}
\label{fig:two}
\end{figure*}

\section{ETRIDriving Dataset}
We introduce ETRIDriving, an autonomous driving dataset labeled with the nine high-level driving actions of the ego-vehicle. To assemble ETRIDriving, we collected data from the sensors (two front-facing cameras, GPS/IMU, CAN bus, and Lidar scanner) mounted on the vehicle (Genesis G80, Hyundai) while driving for approximately 21 hours in various conditions and synchronized the data in time. Next, three annotators, who had driven the vehicle, manually assigned one of the nine actions to every frame in the data. For example, for the data collected when the ego-vehicle was turning left, they assigned the $\textit{Turn left}$ label to the data. Figure \ref{fig:one} shows an example of the labeling process. 

The annotators were instructed to refer to the information (e.g., front-view image, turn signal, steering angle, vehicle trajectory) available from the sensor data according to our guidelines when determining the start and end frame indices for an action. They were also instructed to assign \textit{Unable to define} to the frames when they are unable to find a suitable action. The frames with \textit{Unable to define} were not used for the training and test. The detailed information can be found at https://github.com/d1024choi/ETRIdriving-DevKit. We also open ETRIDriving to the public at the web-page.

\section{Proposed Model}
\subsection{Problem Definition}
Let $\mathbf{P}_{a}=[\mathbf{p}_{1},...,\mathbf{p}_{L}]$ denote a path that matches a driving intention, $a$, determined by a planning system of an AV. Here $\mathbf{p}_{l} \in \mathbf{R}^{2}$ is a position vector in an egocentric coordinate system centered on the AV, and $a$ is an element of the driving-action space, $\mathcal{A}$. We define $\mathcal{A}$ based on the nine actions described in sec. 1. Furthermore, let $\mathbf{I}_{t}$ and $s$, respectively, denote sensor data obtained at the current time, $t$, and a target speed of the AV. Then, our target is to generate $K$ plausible paths, $\{ \mathbf{P}_{a}^{k} \}_{k=1}^{K}$, that match $a$ and $s$ from $\mathbf{I}_{t}$. In the remainder of this paper, we omit time index $t$ for readability. 
% OK, 200603 -----

\subsection{Model Architecture}
The proposed model consists of two networks: the feature extraction network (FEN) and path generation network (PGN). The FEN extracts meaningful features from sensor data (images or point clouds) and the PGN generates paths from the features based on the driving intention and target speed. To make the generated paths plausible and consistent with the intention, we devised the discrimination network (DN) that, through the attention operation, classifies the intention of the paths and distinguishes between the ground-truth and generated paths. We also devised an interaction model between the positions in the paths and the intentions hidden in the positions. The proposed PGN architecture that reflects the interaction model results in the improvement of the accuracy and diversity of the generated paths.
% OK, 200603 -----

\textbf{FEN} Various types of sensors, whose outputs provide the contextual cues of the surroundings, can be considered for inputs to the FEN. In this paper, we used images obtained from a front-facing camera. Consequently, $\mathbf{I}$ denotes the image from the camera at $t$. As the FEN, we used the convolutional layers of $\texttt{ResNet50}$~\cite{He}. The output of the FEN is then visual context vectors $\mathbf{V}=[\mathbf{v}_{1},...,\mathbf{v}_{M}]$, with $M$ denoting the spatial resolution of the output. Our model implicitly model the contextual cues through the visual context vectors. However, one can consider explicitly modeling the cues by using semantic segmentation results of $\mathbf{I}$ as additional inputs to the FEN~ \cite{Kim}.

\textbf{PGN} In sec. 4.1, we define a path as a series of position vectors. In this paper, it is assumed that, when determining a path that matches a driving intention, human drivers sequentially locate each position according to the corresponding intention. For example, to make a right turn, human drivers determine a path in advance a few meters away from the corner as follows: 1) maintain or decrease the speed (a series of $\textit{Go}$), 2) make a right turn (a series of $\textit{Turn Right}$), and 3) settle in a target lane (a series of $\textit{Go}$). Based on this assumption, we devised an interaction model between the positions and intentions hidden in the positions as follows: 
\begin{equation}
a_{l} = f(a_{1:l-1}, \mathbf{p}_{1:l-1} | \mathbf{V}, a)  \label{eqn:one}
\end{equation}
\begin{equation}
\mathbf{p}_{l} = g(\mathbf{p}_{1:l-1}, a_{l} | \mathbf{V}, a),  \label{eqn:two}
\end{equation}
where $f$ and $g$ are mapping functions and $a_{l} \in \mathcal{A}$ denotes the hidden intention for $\mathbf{p}_{l}$. For clarity, we will refer to $a$ and $a_{l}$ as \textit{global intention} (GI) and \textit{local intention} (LI), respectively.

Fig.~\ref{fig:two}-(c) presents a brief description of the proposed PGN structure reflecting the interaction model. It consists of two long short-term memory (LSTM) networks: one for path generation ($LSTM_{P}$) and the other for intention estimation ($LSTM_{A}$). Let $\textbf{a}_{l} \in \mathbf{R}^{|\mathcal{A}|}$ denote a vector representation of $a_{l}$. $\mathbf{p}_{l}$ is then determined using $\textbf{a}_{l}$ via $LSTM_{P}$ as follows:
\begin{equation}
\mathbf{e}_{l-1}^{p} = \phi_{Relu}([\mathbf{p}_{l-1};\mathbf{a}_{l}]), \label{eqn:three}
\end{equation}
\begin{equation}
\mathbf{h}_{l}^{p} = LSTM_{P}(\mathbf{e}_{l-1}^{p}, \mathbf{h}_{l-1}^{p}), \label{eqn:four}
\end{equation}
\begin{equation}
\mathbf{c}_{l}=ATT_{P}(\mathbf{h}_{l}^{p}, \mathbf{V}), \label{eqn:five}
\end{equation}
\begin{equation}
\mathbf{p}_{l} = \phi(\phi_{Relu}([\mathbf{c}_{l};\mathbf{h}_{l}^{p}])) \label{eqn:six},
\end{equation}
where $[;]$ and $ATT_{P}$ denote the concatenation and spatial attention operation~\cite{Lu}, respectively, and $\phi$ and $\phi_{Relu}$, respectively, denote fully-connected layers with linear and Relu activation functions. We initialized the hidden state of $LSTM_{P}$ as follows:
\begin{equation}
\mathbf{h}_{0}^{p} = \phi_{Relu}([\phi_{Relu}(\bar{\mathbf{a}});\mathbf{z}])\label{eqn:seven},
\end{equation}
where $\bar{\mathbf{a}}$ is a one-hot vector representation of $a$, $\mathbf{z}$ is a random noise vector drawn from a zero-mean Gaussian distribution with a standard deviation of 1. Furthermore, we used speed $s$ for the initial position, $\mathbf{p}_{0}=[s, s]^{T}$. 

Hidden intention $a_{l}$ is determined as follows:
\begin{equation}
\mathbf{e}_{l-1}^{a} = \phi_{Relu}([\mathbf{a}_{l-1};\mathbf{p}_{l-1};\mathbf{h}_{l-1}^{p};\mathbf{c}_{l-1}]), \label{eqn:eight}
\end{equation}
\begin{equation}
\mathbf{h}_{l}^{a} = LSTM_{A}(\mathbf{e}_{l-1}^{a}, \mathbf{h}_{l-1}^{a}), \label{eqn:nine}
\end{equation}
\begin{equation}
\mathbf{a}_{l} = \phi(\mathbf{h}_{l}^{a}) \label{eqn:ten}.
\end{equation}
We used $\bar{\mathbf{a}}$ for $\mathbf{a}_{0}$ and a zero vector for the initial hidden state of $LSTM_{A}$, $\mathbf{h}_{0}^{a}$. As shown in Eqn. \ref{eqn:six} and Eqn. \ref{eqn:eight}, the context vector $\mathbf{c}_{l}$ is utilized both for the calculation of $\mathbf{p}_{l}$ and the estimation of $a_{l+1}$. The sharing of $\mathbf{c}_{l}$ allows the PGN to learn how to better utilize $\mathbf{V}$ for the path generation, as verified in sec. 5.

\textbf{DN} DN consists of two LSTM networks: one for path discrimination and classification ($LSTM_{D_{1}}$) and the other for discriminating sequential LIs ($LSTM_{D_{2}}$). The DN takes $\mathbf{P}_{a}$ and $\mathbf{V}$ as its input and outputs $d_{scr1} \in [0, 1]$ and $\mathbf{c}_{cls} \in \mathbf{R}^{|\mathcal{A}|}$ as follows.
\begin{equation}
\mathbf{e}_{l-1}^{d_{1}} = \phi_{Relu}(\mathbf{p}_{l-1}), \label{eqn:eleven}
\end{equation}
\begin{equation}
\mathbf{h}_{l}^{d_{1}} = LSTM_{D_{1}}(\mathbf{e}_{l-1}^{d_{1}}, \mathbf{h}_{l-1}^{d_{1}}), \label{eqn:twelve}
\end{equation}
\begin{equation}
\mathbf{c}_{l}=ATT_{D}(\mathbf{h}_{l}^{d_{1}}, \mathbf{V}), \label{eqn:thirteen}
\end{equation}
\begin{equation}
d_{scr1} = \phi_{sig}(\sum_{l=1}^{L}\mathbf{h}_{l}^{d_{1}}) \label{eqn:fourteen},
\end{equation}
\begin{equation}
\mathbf{c}_{cls} = \phi(\sum_{l=1}^{L}\phi_{Relu}[\mathbf{h}_{l}^{d_{1}};\mathbf{c}_{l}]) \label{eqn:fifteen},
\end{equation}
where $\phi_{sig}$ and $ATT_{D}$ denote a fully-connected layer with the sigmoid activation function and spatial attention operation~\cite{Lu}, respectively. 

The DN also takes a sequence of LI vectors, $\mathbf{A} = \{ \bar{\mathbf{a}}_{l} \}_{l=1}^{L}$, as an input and outputs $d_{scr2} \in [0, 1]$ as follows.
\begin{equation}
\mathbf{e}_{l-1}^{d_{2}} = \phi_{Relu}(\bar{\mathbf{a}}_{l-1}), \label{eqn:sixteen}
\end{equation}
\begin{equation}
\mathbf{h}_{l}^{d_{2}} = LSTM_{D_{2}}(\mathbf{e}_{l-1}^{d_{2}}, \mathbf{h}_{l-1}^{d_{2}}), \label{eqn:seventeen}
\end{equation}
\begin{equation}
d_{scr2} = \phi_{sig}(\mathbf{h}_{L}^{d_{2}}) \label{eqn:eighteen},
\end{equation}
where $\bar{\mathbf{a}}_{l}$ is the one-hot vector representation of $a_{l}$. We used zero vectors for the initial hidden states, $\mathbf{h}_{0}^{d_{1}}$ and $\mathbf{h}_{0}^{d_{2}}$. 
% OK again, 200603 -----

$d_{scr1}$ indicates if the input path is real or fake. As shown in Eqn.~\ref{eqn:fourteen}, $d_{scr1}$ is determined by $\{ \mathbf{h}_{l}^{d} \}$, which are calculated based on the positions only. This implies that the DN only considers the shape of the path in path discrimination. Further, $\mathbf{c}_{cls}$ indicates the category of driving intention under which the input path is subsumed. When calculating $\mathbf{c}_{cls}$, the DN considers both the input path and visual contexts simultaneously through the attention mechanism, as expressed in Eqn.~\ref{eqn:fifteen}. Finally, $d_{scr2}$ indicates if the sequence of LIs is real or fake. Consequently, the PGN, guided by the DN through the adversarial training, produces paths that are realistic and consistent with the input driving intentions. 
% OK again, 200603 -----

\subsection{Losses}
\textbf{Variety Loss} To enable the PGN to produce diverse paths, we used the variety loss~\cite{Gupta, Thiede}, which is defined as follows.
\begin{equation}
\mathcal{L}_{var} = \min_{k} \frac{1}{L}||\mathbf{P}_{a}^{gt} - \mathbf{P}_{a}^{k}||_{2}^{2},  \label{eqn:nineteen}
\end{equation}
where $\mathbf{P}_{a}^{gt}$ is the ground-truth path from the dataset. 
% Check again, 200603 -----

\textbf{Adversarial Losses} For simplicity, let $c$ denote the conditional inputs, $( \mathbf{I}, a, s )$. We defined two adversarial losses, one for path $\mathbf{P}$ and the other for a sequence of LI vectors $\mathbf{A}$, as follows.
\begin{multline}
\mathcal{L}_{adv1} = \mathbb{E}_{\mathbf{P}\sim p_{data}}[\mathtt{log} D_{1}(\mathbf{P})] \\ 
+ \mathbb{E}_{\mathbf{z}\sim p_{z}}[\mathtt{log} (1-D_{1}(G_{p}(\mathbf{z}|c)))],  \label{eqn:twenty}
\end{multline}
\begin{multline}
\mathcal{L}_{adv2} = \mathbb{E}_{\mathbf{A}\sim p_{data}}[\mathtt{log} D_{2}(\mathbf{A})] \\ 
+ \mathbb{E}_{\mathbf{z}\sim p_{z}}[\mathtt{log} (1-D_{2}(G_{a}(\mathbf{z}|c)))],  \label{eqn:twentyone}
\end{multline}
where $D_{1}$ and $D_{2}$, respectively, denote the path and sequence DNs. In addition, $G_{p}$ and $G_{a}$ denote the generation networks for the path and sequence, respectively. 
% Check again, 200603 -----

\textbf{Classification Losses} We defined two classification losses, one for $a$ and the other for $\{a_{l}\}$, as follows.
\begin{equation}
\mathcal{L}_{cls1} = \texttt{BCE}(\bar{\mathbf{a}}, \texttt{softmax}(\mathbf{c}_{cls})),  \label{eqn:twentytwo}
\end{equation}
\begin{equation}
\mathcal{L}_{cls2} = \frac{1}{L-1} \sum_{l=2}^{L} \texttt{BCE}(\bar{\mathbf{a}}_{l}, \texttt{softmax}(\mathbf{a}_{l})), \label{eqn:twentythree}
\end{equation}
where $\bar{\mathbf{a}}$ and $\bar{\mathbf{a}}_{l}$ respectively denote the one-hot vector representations of $a$ and $a_{l}$. 
% Check again, 200603 -----

\textbf{Full Objective} The final objective functions to be minimized are as follows.
\begin{equation}
\mathcal{L}_{G} = \lambda_{1}\mathcal{L}_{var} + \mathcal{L}_{adv1} + \mathcal{L}_{cls1} + \lambda_{2}\mathcal{L}_{adv2} + \lambda_{3} \mathcal{L}_{cls2},  \label{eqn:twentyfour}
\end{equation}
\begin{equation}
\mathcal{L}_{D} = -\mathcal{L}_{adv1} -\lambda_{4}\mathcal{L}_{adv2} + \mathcal{L}_{cls1},  \label{eqn:twentyfive}
\end{equation}
where we set $\lambda_{1}=10^{2}$, $\lambda_{2}=10^{-2}$, $\lambda_{3}=5 \times 10^{-3}$, and $\lambda_{4}=1$ for our experiments. 
% Check again, 200603 -----

\section{Experiments}
\subsection{Evaluation Methodology}
There are 129 sequences of 10-minutes duration in ETRIDriving. For evaluation, we selected 32 sequences (25 for training and seven for testing) with diverse driving actions of the ego-vehicle. This is because, in most sequences, the vehicle maintains a straight trajectory or stands still most of the time. The samples, $(\mathbf{I}, \mathbf{P}_{a}, a, \mathbf{A}, s)$, for training and testing were generated from the 32 sequences as follows. First, a transformation matrix, $\mathbf{M}_{t}$, was created from the yaw, pitch, roll, and global position of the vehicle at current time $t$. We used $\mathbf{M}_{t}$ to represent the future trajectory of the vehicle in the egocentric coordinate system. Each position in $\mathbf{P}_{a}$ was then obtained by using the positions in the transformed trajectory under constraints $||\mathbf{p}_{l} - \mathbf{p}_{l+1} ||^2 = 1$, $\mathbf{p}_{0}=[0,~0]^{T}$. We set $L=20$, which indicated that the length of a path was 20 meters. For $a_{l} \in \mathbf{A}$, we found the position in the trajectory nearest to $\mathbf{p}_{l}$ and used the driving-action label assigned to the position. For $a$, we chose one action from $\{a_{l}\}_{l=1}^{F}$, where $F \leq L$. However, it was not easy to find the one that matched the ``true'' driving intention of $\mathbf{P}_{a}$, when there were several types of actions in $\{a_{l}\}_{l=1}^{F}$. To address this problem, we randomly selected $a_{l}$ from $\{a_{l}\}_{l=1}^{F}$ during the training, such that there was a high probability of choosing the action constituting the majority of $\{a_{l}\}_{l=1}^{F}$ for $a$. In contrast, during the test, we chose the action constituting the majority of $\{a_{l}\}_{l=1}^{F}$ for consistency. Finally, the front-facing camera image and the speed of the vehicle recorded at $t$ were used for $\mathbf{I}$ and $s$, respectively. 

It was possible to obtain approximately 162,000 samples from the 32 sequences. However, we discarded samples from trajectories obtained under low GPS accuracy. We also discarded some samples to balance the numbers of the different actions in the dataset. Consequently, 28,320 samples (21,876 for the training and 6,444 for the testing) were finally used. 

% OK, 200608 ----
% ------------------------------------------------------------ Table 1
\begin{table*}
\footnotesize
\begin{center}
\begin{tabular}{|l|c|c|c|c|c|}
\hline
Method & minADE($\downarrow$) & minFDE($\downarrow$) & Div($\uparrow$) & MLL($\uparrow$) & minMSE-S($\downarrow$) \\
\hline\hline
E2ENet-BR~\cite{Codevilla} & - & - & - & - & $0.0052\pm0.0002$ \\
VTGNet~\cite{Cai} & $0.351\pm0.020$ & $0.998\pm0.054$ & - & - & $0.0064\pm0.0005$ \\
PathGen-single & $0.339\pm0.002$ & $0.868\pm0.022$ & - & - & $0.0049\pm0.0005$ \\
CarNet~\cite{Sadeghian2} & $0.258\pm0.012$ & $0.733 \pm 0.027$ & - & - & $\textcolor{red}{0.0029}\pm\textcolor{red}{0.0005}$ \\
R2P2-GC~\cite{Rhinehart2} & $\textcolor{blue}{0.135}\pm\textcolor{blue}{0.019}$ & $\textcolor{red}{0.197}\pm\textcolor{red}{0.023}$ & - & - & $0.0051\pm0.0006$ \\
CVAE~\cite{Lee} & $0.164\pm0.012$ & $0.293\pm0.005$ & $\textcolor{red}{1.405}\pm\textcolor{red}{0.599}$ & $-2.235\pm0.144$ & $0.0125\pm0.0034$ \\

\hline
PathGAN-1 (P1) & $0.284\pm0.018$ & $0.637\pm0.043$ & $0.258\pm0.019$ & $-2.557\pm0.059$ & $0.0064\pm0.0007$ \\
% PathGAN-1 (P1) & $0.161\pm0.013$ & $0.336\pm0.041$ & $0.389\pm0.059$ & $-2.168\pm0.032$ & $0.0043\pm0.0004$ \\
PathGAN-2 (P2) & $0.144\pm0.004$ & $0.304\pm0.015$ & $0.340\pm0.004$ & $\textcolor{green}{-2.106}\pm\textcolor{green}{0.006}$ & $\textcolor{blue}{0.0036}\pm\textcolor{blue}{0.0002}$ \\

PathGAN-3 (P3) & $\textcolor{green}{0.140}\pm\textcolor{green}{0.005}$ & $\textcolor{blue}{0.270}\pm\textcolor{blue}{0.014}$ & $\textcolor{blue}{0.422}\pm\textcolor{blue}{0.048}$ & $\textcolor{red}{-2.098}\pm\textcolor{red}{0.011}$ & $0.0047\pm0.0009$ \\

PathGAN-4 (P4) & $\textcolor{red}{0.134}\pm\textcolor{red}{0.009}$ & $\textcolor{green}{0.271}\pm\textcolor{green}{0.023}$ & $\textcolor{green}{0.421}\pm\textcolor{green}{0.072}$ & $\textcolor{blue}{-2.101}\pm\textcolor{blue}{0.003}$ & $\textcolor{green}{0.0040}\pm\textcolor{green}{0.0004}$ \\

\hline
\end{tabular}
\end{center}
\caption{Quantitative results of all models when $K=20$ and $F=5$. minADE and minFDE respectively denote the minimum ADE and FDE among the $K$ paths. minMSE-S is calculated using the path with the minimum ADE. The unit of ADE and FDE is meter. To obtain the values in the table, we trained each model three times. For the sake of readability, the best, second-best, and third-best results in each metric are displayed in red, blue, and green, respectively.}
\label{table:one}
\end{table*}

% ------------------------------------------------------------ Table 2
\begin{table}
\scriptsize
\begin{center}
\begin{tabular}{|c|c|c|c|}
\hline
Model & Path Disc./Class. & Interaction Model & Seq. LIs Disc. \\
\hline\hline
% P0 & \xmark & \xmark & \xmark & \xmark\\
P1 & \xmark & \xmark & \xmark\\
P2 & \cmark & \xmark & \xmark\\
P3 & \cmark & \cmark & \xmark\\
P4 & \cmark & \cmark & \cmark\\
\hline
\end{tabular}
\end{center}
\caption{Ablation study on the effectiveness of adversarial training and interaction model.}
\label{table:two}
\end{table}

% ------------------------------------------------------------ Table 3
\begin{table}
\scriptsize
\begin{center}
\begin{tabular}{|c|c|c|c|c|}
\hline
Model & minADE($\downarrow$) & minFDE($\downarrow$) & Div($\uparrow$) & MLL($\uparrow$) \\
\hline\hline
P2 & 0.144 & 0.304 & 0.340 & -2.106\\
%P2-A & 0.166 & 0.381 & 0.344 & -2.172 \\
P3-A & 0.149 & 0.310 & 0.335 & -2.114 \\
P3-B & $\textbf{0.140}$ & $\textbf{0.270}$ & $\textbf{0.422}$ & $\textbf{-2.098}$\\
P3-C & 0.152 & 0.302 & 0.397 & -2.118 \\

\hline
\end{tabular}
\end{center}
\caption{Ablation study on the PGN architecture reflecting the interaction model.}
\label{table:three}
\end{table}

\subsection{Baselines}
The proposed model, which we call \textit{PathGAN}, was compared with the following baselines:\\
$\bullet$~$\textbf{PathGen-single}$: Single path generator (FEN and PGN$^{\dagger}$) trained via MSE loss only. Here, PGN$^{\dagger}$ denotes PGN with $LSTM_{A}$ removed.\\
$\bullet$~$\textbf{CVAE}$: Conditional VAE, which models generative distributions conditioned on sensory inputs and a driving intention. Our implementation was based on the deep generative model proposed in~\cite{Lee}.\\
$\bullet$~$\textbf{CarNet}$: Clairvoyant attentive recurrent network proposed in~\cite{Sadeghian2}. We slightly modified the original implementation to make the network take a driving intention as an input.\\
$\bullet$~$\textbf{E2ENet-BR}$: End-to-end model that generates a steering angle from a sensory input, given a driving intention and speed. We followed a branched model proposed in~\cite{Codevilla} for our implementation.\\
$\bullet$~$\textbf{VTGNet}$: End-to-end model that generates a trajectory from sensory inputs~\cite{Cai}. Nine models with the same structure were trained separately for the nine actions.\\
$\bullet$~$\textbf{R2P2-GC}$: Goal-conditioned (GC) deep imitative model proposed in~\cite{Rhinehart2}. Following the original implementation, we first generated $K$ paths using the imitative model. Next, given the goal position, we optimized the proposed cost function to obtain the final planned path. For the goal position, we used the final position of the ground-truth path, which is a strong clue for the path planning.

\subsection{Evaluation Metrics}
We compared the models objectively based on five metrics.\\
$\bullet$~Average Displacement Error (ADE): Average L2 distance between the ground truth and generated paths across all the positions.\\
$\bullet$~Final Displacement Error (FDE): L2 distance between the last ground truth and generated positions.\\
$\bullet$~Diversity (Div): Average of ADE between the generated $K$ paths.\\
$\bullet$~Marginal Log-Likelihood (MLL): Average log-likelihood of the ground truth under the marginalized learned distribution for every position~\cite{Thiede}.\\
$\bullet$~Mean Squared Error for Steering Angle (MSE-S): Squared error between the ground-truth steering angle and its estimate. 

We introduced the MSE-S to evaluate the suitability of a generated path for controlling the real vehicle. To produce a steering angle from a path, we trained an LSTM network that takes a path and vehicle speed as inputs and outputs a steering angle. Note that the angles were scaled to be in range [-1, 1] for the training.

\subsection{Implementation Details}
The FEN produces a feature map of size 20$\times$10$\times$2048 from an input image of size 640$\times$320$\times$3 pixels. Therefore, $\mathbf{v}_{i}$ is a vector of length 2048, and the size of $\mathbf{V}$ is 200. The dimension of $\mathbf{v}_{i}$ is reduced to 128 before being fed into the PGN and DN. The dimensions of the hidden state for the $LSTM_{D_{2}}$ and the other $LSTM$ networks were set to 32 and 128, respectively. We trained the PGN and DN simultaneously using Adam~\cite{Kingma} with an initial learning rate of $10^{-4}$ and a batch size of 16 for 100 epochs. The FEN, which was initialized using the trained parameters of the FEN of PathGen-single, was also trained simultaneously using the Adam with an initial learning rate of $5 \times 10^{-5}$. More details can be found in the supplementary material.

\subsection{Quantitative Evaluation}
\subsubsection{Performance Comparison} 
We compared our model based on the five metrics against the baselines. The results are shown in Table \ref{table:one}. As shown in the table, our final model PathGAN-4 (P4) exhibited excellent performance in terms of both the path generation accuracy (ADE, FDE, MLL) and diversity (Div). As expected, CarNet, VTGNet, and PathGen-single underperformed, compared to P4 because they could generate only one path. CVAE also outperformed the three, based on the ADE and FDE, as it could generate multiple paths. However, the MLL and Div results of CVAE indicated that although many of the paths generated by CVAE were diverse, they were inaccurate. R2P2-GC achieved a remarkable performance, especially in terms of the FDE. This is because the final position of the ground-truth path was used as the goal position for the optimization R2P2-GC performed. However, as shown in Table \ref{table:five}, R2P2-GC is approximately 278 times slower than P4 because of its optimization process. It is found from our experiments that when P4 generates 278 paths per test sample (K=278), it achieves the \textbf{minADE 0.08} and \textbf{minFDE 0.124}. On the other hand, in terms of the minMSE-S, CarNet exhibited the best performance. P4 also exhibited a good performance. Specifically, it outperformed E2ENet-BR, which estimated a steering angle from $\mathbf{I}$ directly and has been effective in real-world driving~\cite{Codevilla}.

\subsubsection{Ablation Study} 
To show the effectiveness of the proposed adversarial training and interaction model, we conducted an ablation study on our model in four different settings. Table \ref{table:two} summarizes the differences in the four settings. We can observe from Table \ref{table:one} and~\ref{table:two}  that introducing the path discriminator and classifier improved the accuracy and diversity of the paths generated by P1 significantly (P1 v.s. P2). Furthermore, the performances were further improved by the introduction of the interaction model (P2 v.s. P3). Finally, introducing the sequential LIs discriminator made the paths generated by P3 more suitable for real-world driving (P3 v.s. P4). 

\subsection{Qualitative Evaluation}
\begin{figure*}
\centerline {\includegraphics[width=17.0cm]{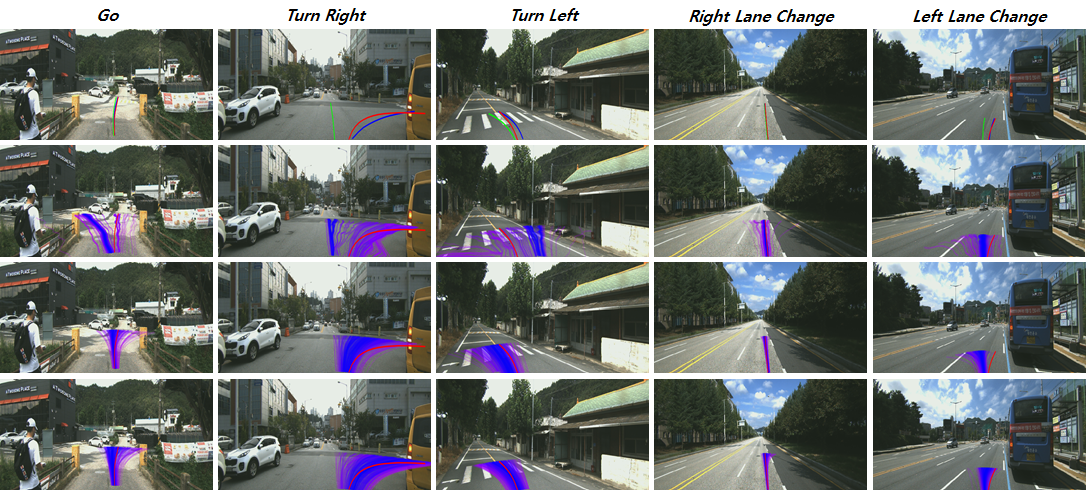}}
\caption{Path generation results. Each row is the results of R2P2-GC, CVAE, P2, and P4 in order. Red lines denote the ground-truth paths. Blue, purple, and green lines denote the generated paths. Blue and purple denote highly probable paths and less probable ones, respectively. We also used green for the paths corresponding to the initial path used in the optimization of R2P2-GC. Finally, 300 paths were generated for each scene from the trained models listed in Table~\ref{table:one}.}
\label{fig:three}
\end{figure*}

In Fig.~\ref{fig:three}, we show the input scene images overlaid with the paths generated by different models. The rows in the figure are the results of R2P2-GC, CVAE, P2, and P4, in order. As can be seen from the figure, the proposed models can generate diverse plausible paths that are consistent with the driving intentions. Furthermore, it can also be observed that R2P2-GC can generate plausible paths from the initial paths, even if the initials were inaccurate. This is because R2P2-GC utilized the final position of the ground-truth as the goal position for its path optimization process. On the other hand, CVAE appeared to fail at generating paths that matched the intentions. As seen in the figure, many of the generated paths were quite different from the ground truth. 

\begin{figure}
\centerline {\includegraphics[width=8.5cm]{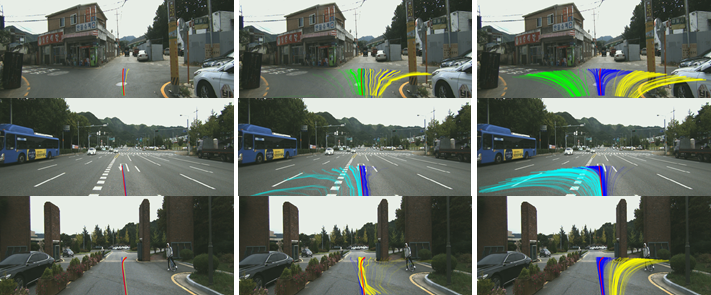}}
\caption{Path generation with various driving intentions. Each column is the result of CarNet, CVAE, and P4 in order. Red lines denote the ground-truth paths. Blue, green, yellow, and sky blue lines denote the paths generated with $\textit{Go}$, $\textit{Turn Left}$, $\textit{Turn Right}$, and $\textit{U-Turn}$, respectively.}
\label{fig:four}
\end{figure}

% generate paths match the driving intention
To test the ability to generate paths corresponding to different driving intentions from a driving scene, we allowed each model generate paths with various driving intentions in driving scenes with possibly more than one intention. The results are shown in Fig.~\ref{fig:four}. The columns in the figure are the results of CarNet, CVAE, and P4, in order. As can be seen in the figure, P4 successfully generated paths that are both plausible and consistent with the driving intentions. In contrast, it appeared that both CarNet and CVAE had challenges understanding the scenes; therefore, they failed to generate paths that matched the intentions. 
% OK, 200608 ----

% figure 3 --------------------------------------------------------------------

% figure 4 --------------------------------------------------------------------

% figure 5 --------------------------------------------------------------------
\begin{figure*}
\centerline {\includegraphics[width=16.7cm, height=7.0cm]{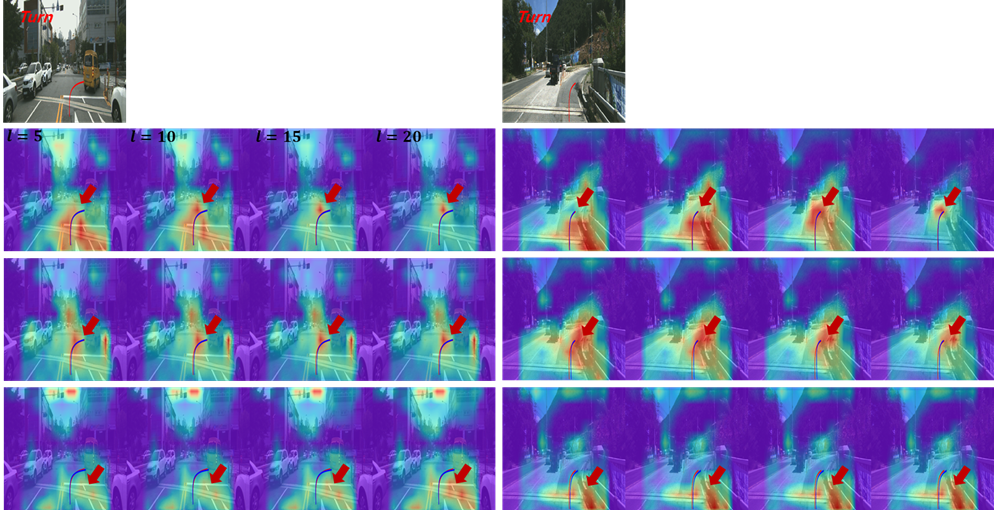}}
\caption{Visualization of attention regions for generating the fifth, tenth, fifteenth, and twentieth positions. The images in the first row are the input scenes. The second, third, and fourth rows are the results of P2, P3-B, and P3-C, respectively.}
\label{fig:five}
\end{figure*}

% figure 6 --------------------------------------------------------------------
\begin{figure}
\centerline {\includegraphics[width=7.5cm, height=3.5cm]{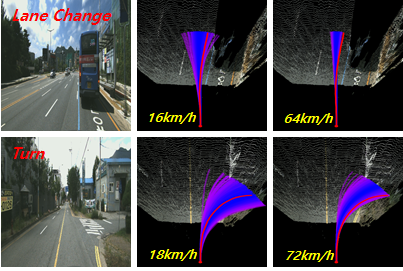}}
\caption{Path generation with different speeds. The images in the first column are the input scenes. The generated paths are depicted in the second and third columns (top-view).}
\label{fig:six}
\end{figure}

% figure 8 --------------------------------------------------------------------
% \begin{figure}
% \centerline {\includegraphics[width=8.0cm]{fig8_new.png}}
%\caption{t-SNE~\cite{Maaten} plots of $[\mathbf{a}_{1};\mathbf{a}_{2};...;\mathbf{a}_{L}]$. The left is the result from the ground-truth while the middle and right are from P3 and P4, respectively.}
%\label{fig:eight}
%\end{figure}

\subsection{Analysis}

\subsubsection{Network Architecture for Interaction Model}
To show the effectiveness of the proposed PGN architecture depicted in Fig.~\ref{fig:two}-(c), we trained PathGAN-3 (P3) with different PGN architectures. Table~\ref{table:three} shows the results. In the table, P2 denotes the PGN architecture of PathGAN-2 depicted in Fig.~\ref{fig:two}-(a). P3-A, P3-B, and P3-C respectively denote P3 with the PGN of the architecture depicted in Fig.~\ref{fig:two}-(b), (c), and (d). In P3-A, the LI $\mathbf{a}_{l+1}$ for the position $\mathbf{p}_{l+1}$ is directly estimated from the context vector $\mathbf{c}_{l}$ through a fully-connected network. In P3-B, the visual attention operation is performed once for the calculation of $\mathbf{c}_{l}$, which is utilized both for the calculation of $\mathbf{p}_{l}$ and for the estimation of $\mathbf{a}_{l+1}$. Note here that P3-B is the same as P3. Finally, in P3-C, a separate visual attention operation is performed for $\mathbf{p}_{l}$ and $\mathbf{a}_{l+1}$, respectively. We can observe three facts from the table. First, the estimation of LIs without taking the interaction into account does not help improve the performance (P3-A v.s. P2). Second, the proposed interaction model improves the diversity of the generated paths (P3-B/C v.s. P2). Third, the PGN architecture that calculates $\mathbf{c}_{l}$ once for both $\mathbf{p}_{l}$ and $\mathbf{a}_{l+1}$ is more effective in improving the performances than the architecture that calculates separate $\mathbf{c}_{l}$ for $\mathbf{p}_{l}$ and $\mathbf{a}_{l+1}$, respectively (P3-B v.s. P3-C).

In Fig.~\ref{fig:five}, we highlighted the image regions the PGN paid attention to during the path generation. The second, third, and fourth rows in the figure are the results of P2, P3-C, and P3-D, respectively. We can see in the figure that P2 paid attention to different regions for each position. In contrast, P3-C and P3-D, from the beginning of the generation process, paid attention to the regions required to generate all the positions and changed their interests, as the process progressed. It seems that the PGNs were trained to utilize non-local visual features owing to the proposed interaction model. Similar tendencies can be observed in other examples in the test dataset.

\subsubsection{Effect of $F$} 
As we mentioned in section 5.1, during the training, we randomly selected one driving intention among $F$ candidates that corresponded to the $F$ positions closest to the ego-vehicle. To find the best $F$, we trained our models using different $F$. Table~\ref{table:four} shows the ADE and FDE performances of P2 and P4 with different $F$. We can observe from the table that $F=5$ yielded the best performances. The best $F$ value may be closely related to the average speed of the vehicle during the data collection because the look-ahead distance (the distance between the vehicle and a point in front of the vehicle that a human driver may look at to track the roadway) is known to have a positive correlation with the speed of the vehicle. In ETRIDriving, the average speed of the ego-vehicle was approximately 32 km/h.

\subsubsection{Effect of Speed} 
To investigate the role of the input speed during the path generation process, we generated paths on different speeds. Figure~\ref{fig:six} shows the results of P4. The second and third columns of the figure are the results for the low and high speeds, respectively. As can be seen in the figure, our model generated high-curvature paths on low speeds and low-curvature paths on high speeds. This tendency is similar to the human driving pattern wherein the steering angle is gradually changed when driving on high speed.

\subsubsection{Path Generation Speed} 
We show the path generation speeds of the models in Table~\ref{table:five}. The values in the table were obtained by averaging 300 speed records. To obtain the records, we ran the models on a PC equipped with Intel i7 CPU (@4.0GHz), 16 GB RAM, GTX 1080Ti GPU. The table shows that the proposed model is comparable to CVAE and faster than the others. R2P2-GC had the worst performance (approximately 278 times slower than P4) mainly due to its optimization process.

% ------------------------------------------------------------ Table 4
\begin{table}
\scriptsize
\begin{center}
\begin{tabular}{|c|c|c|c|c|}
\hline
Model & $F=1$ & $F=5$ & $F=10$ & $F=20$ \\
\hline\hline
P2 & 0.189/0.414 & $\mathbf{0.144}$/$\mathbf{0.304}$ & 0.146/0.289 & 0.182/0.362\\
P4 & 0.142/0.290 & $\mathbf{0.134}$/$\mathbf{0.271}$ & 0.145/0.275 & 0.164/0.332\\
\hline
\end{tabular}
\end{center}
\caption{minADE/minFDE performances with $K=20$ and varying $F$.}
\label{table:four}
\end{table}

% Table 5 --------------------------------------------------------
\begin{table}[t]
\scriptsize
\centering
  \begin{tabular}{|l|c|c|c|c|c|c|c|}
    \hline
    Model & VTGNet     & CarNet  & CVAE  & R2P2-GC & P4 \\ \hline 
    Time & 0.062 & 0.037 & 0.010 & 5.012  & 0.018 \\ \hline 
  \end{tabular}
\caption{Path generation speed comparison. The unit of the values is second per one path generation.}  
\label{table:five}
\end{table}

\section{Conclusion}
In this paper, we proposed a model that can generate multiple paths from egocentric images, given a driving intention and speed. To generate plausible paths that are consistent with the driving intention, we utilized a GAN framework wherein the path discrimination and classification networks were simultaneously trained with the generative model. The accuracy and diversity of the generated paths were further enhanced by the proposed interaction model for the positions in a path and the intentions hidden in them. To evaluate the proposed model, we introduced ETRIDriving, a driving dataset labeled with the nine high-level driving actions, and verified the state-of-the-art performance of the model on ETRIDriving in terms of the accuracy and diversity.

%{
%\bibliographystyle{ieee}
%\bibliography{egbib}
%}

\end{document}